%% file: 00_preamble.tex
\newcommand{\ignore}[1]{}

\documentclass[10pt,twocolumn,letterpaper]{article}

\usepackage{iccv}
\usepackage{times}
\usepackage{epsfig}
\usepackage{graphicx}
\usepackage{amsmath}
\usepackage{amssymb}
\usepackage{booktabs}

\DeclareMathOperator*{\argmax}{arg\,max}

\usepackage[breaklinks=true,bookmarks=false]{hyperref}

\hypersetup{
   colorlinks=true
}

\iccvfinalcopy 

\def\iccvPaperID{2616} 

\begin{document}

\title{Submodular Trajectory Optimization for Aerial 3D Scanning}

\author{
Mike Roberts$^{1,2}$~~~~Debadeepta Dey$^{2}$~~~~Anh Truong$^{3}$~~~~Sudipta Sinha$^{2}$\\
Shital Shah$^{2}$~~~~Ashish Kapoor$^{2}$~~~~Pat Hanrahan$^{1}$~~~~Neel Joshi$^{2}$\\
\vspace{-5pt}\\
$^{1}$Stanford University~~~~$^{2}$Microsoft Research~~~~$^{3}$Adobe Research
}

\maketitle


\begin{abstract}
Drones equipped with cameras are emerging as a powerful tool for large-scale aerial 3D scanning, 
but existing automatic flight planners do not exploit all available information about the scene, and can therefore produce inaccurate and incomplete 3D models. 
We present an automatic method to generate drone trajectories, such that the imagery acquired during the flight will later produce a high-fidelity 3D model. Our method uses a coarse estimate of the scene geometry to plan camera trajectories that: (1) cover the scene as thoroughly as possible; (2) encourage observations of scene geometry from a diverse set of viewing angles; (3) avoid obstacles; and (4) respect a user-specified flight time
budget. Our method relies on a mathematical model of scene coverage that exhibits an intuitive diminishing returns property known as submodularity.
We leverage this property extensively to design a trajectory planning algorithm  that reasons globally about the non-additive coverage reward obtained across a trajectory, jointly with the cost of traveling between views.
We evaluate our method by using it to scan three large outdoor scenes, and we perform a quantitative evaluation using a photorealistic video game simulator.
\end{abstract}

\input{10_intro.tex}
\input{20_related_work.tex}
\input{30_approach.tex}

\input{40_results.tex}

\input{50_discussion.tex}

{\small
\bibliographystyle{ieee}
\bibliography{00_preamble}
}

\end{document}

%% file: 10_intro.tex
\begin{figure}[t]
\begin{center}
\includegraphics[width=0.47\textwidth]{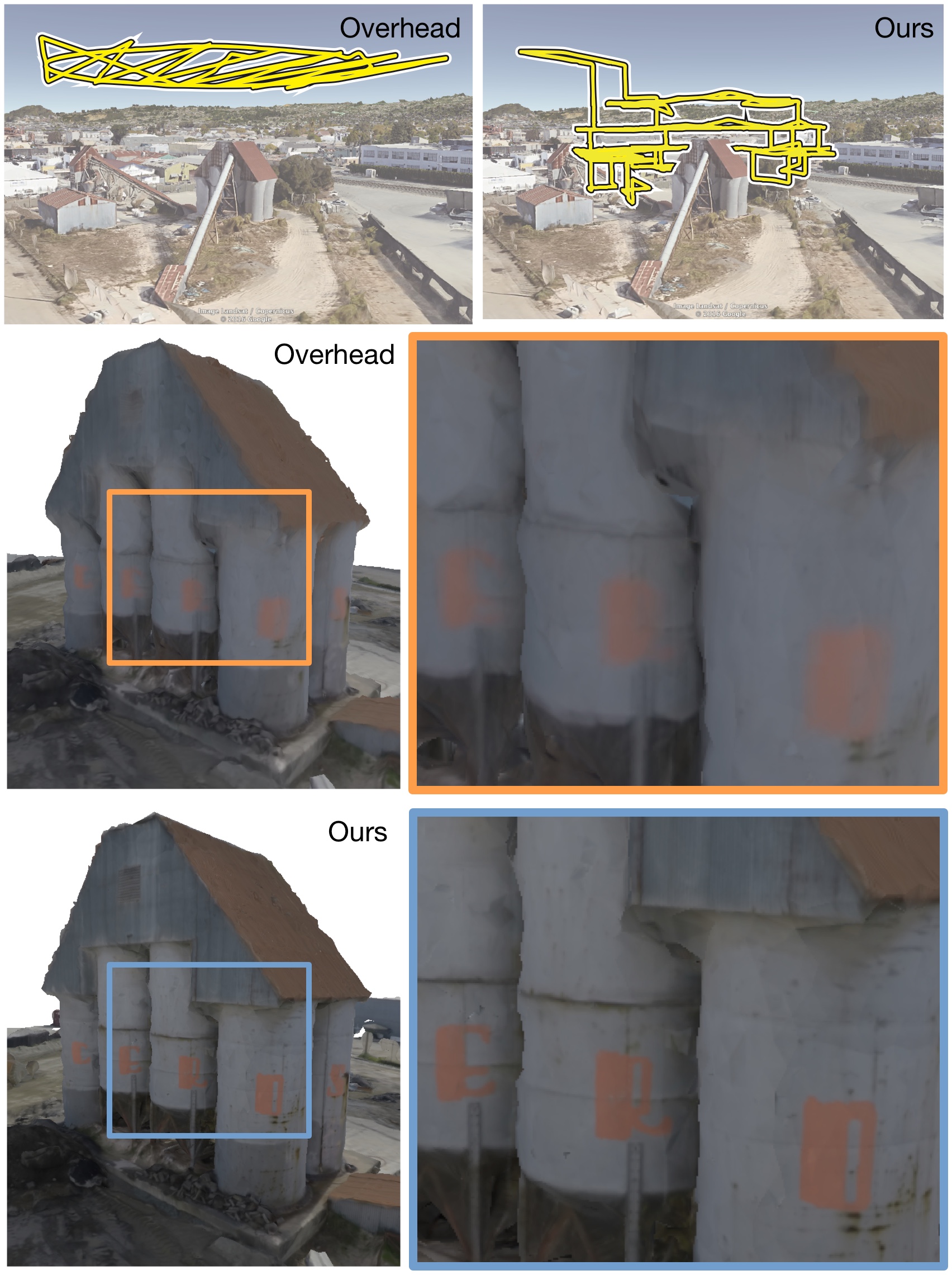}{\vspace{-7pt}}
\end{center}
\caption{
3D reconstruction results obtained using our algorithm for generating aerial 3D scanning trajectories, as compared to an overhead trajectory.
Top row: Google Earth visualizations of the trajectories.
Middle and bottom rows: results obtained by flying a drone along each trajectory, capturing images, and feeding the images to multi-view stereo software.
Our trajectories lead to noticeably more detailed 3D reconstructions than overhead trajectories.
In all our experiments, we control for the flight time, battery consumption, number of images, and quality settings used in the 3D reconstruction.
}
\vspace{-13pt}
\label{fig:teaser}
\end{figure}

\vspace{-16pt}
\section{Introduction}

Small consumer drones equipped with high-resolution cameras are emerging as a powerful tool for large-scale aerial 3D scanning.
In order to obtain high-quality 3D\ reconstructions, a drone must capture images that densely cover the scene.
Additionally, 3D reconstruction methods typically require surfaces to be viewed from multiple viewpoints, at an appropriate distance, and with sufficient angular separation (i.e., baseline) between views.
Existing autonomous flight planners do not always satisfy these  requirements, which can be difficult to reason about, even for a skilled human pilot manually controlling a drone.
Furthermore, the limited battery life of consumer drones provides only 10--15 minutes of flight time, making it even more challenging to obtain high-quality 3D reconstructions.


In lieu of manual piloting, commercial flight planning tools generate conservative trajectories (e.g., a lawnmower or orbit pattern at a safe height above the scene) that attempt to cover the scene while respecting flight time budgets \cite{3dr:2017a,pix4d:2017a}.
However, because these trajectories are generated with no awareness of the scene geometry, they tend to over-sample some regions (e.g., rooftops), while under-sampling others (e.g., facades, overhangs, and fine details), and therefore sacrifice reconstruction quality.

We propose a method to automate aerial 3D scanning, by planning good camera trajectories for reconstructing large 3D scenes (see Figure \ref{fig:teaser}).
Our method relies on a mathematical model that evaluates the usefulness of a camera trajectory for the purpose of 3D scanning.
Given a coarse estimate of the scene geometry as input, our model quantifies how well a trajectory covers the scene, and also quantifies the diversity and appropriateness of views along the trajectory.
Using this model for scene coverage, our method generates trajectories that maximize coverage, subject to a travel budget.
We bootstrap our method using coarse scene geometry, which we obtain using the imagery acquired from a short initial flight over the scene. 



We formulate our trajectory planning task as a reward-collecting graph optimization problem known as \textit{orienteering}, that combines aspects of the traveling salesman and knapsack problems, and is known to be NP-hard \cite{gunawan:2016,vansteenwegena:2011}.
However, unlike the additive rewards in the standard orienteering problem, our rewards are non-additive, and globally coupled through our coverage model.
We make the observation that our coverage model exhibits an intuitive diminishing returns property known as \textit{submodularity}~\cite{krause:2014}, and therefore we must solve a \textit{submodular orienteering} problem.
Although submodular orienteering is strictly harder than additive orienteering, it exhibits useful structure that can be exploited.
We propose a novel transformation of our submodular orienteering problem into an additive orienteering problem, and we solve the additive problem as an integer linear program. We leverage submodularity extensively throughout the derivation of our method, to obtain approximate solutions with strong theoretical guarantees, and dramatically reduce computation times.

We demonstrate the utility of our method by using it to scan three large outdoor scenes: a barn, an office building, and an industrial site.
We also quantitatively evaluate our algorithm in a photorealistic video game simulator.
In all our experiments, we obtain noticeably higher-quality 3D reconstructions than strong baseline methods.
%



%% file: 20_related_work.tex
\vspace{-5pt}
\section{Related Work}

\paragraph{Aerial 3D Scanning and Mapping}
High-quality 3D reconstructions of very large scenes can be obtained using offline multi-view stereo algorithms \cite{furukawa:2015} to process images acquired by drones \cite{pix4d:2015}.
Real-time mapping algorithms for drones have also been proposed, that take as input either RGBD \cite{heng:2011,loianno:2015,michael:2012,sturm:2013} or RGB \cite{wendel:2012} images, and produce as output a 3D\ reconstruction of the scene.
These methods are solving a reconstruction problem, and do not, themselves, generate drone trajectories.
Several commercially available flight planning tools have been developed to assist with 3D scanning \cite{3dr:2017a,pix4d:2017a}.
However, these tools only generate conservative lawnmower and orbit trajectories above the scene.
In contrast, our algorithm generates trajectories that cover the scene as thoroughly as possible, ultimately leading to higher-quality 3D reconstructions.

Generating trajectories that \emph{explore} an unknown environment, while building a map of it, is a classical problem in robotics \cite{thrun:2005}.
Exploration algorithms have been proposed for drones based on local search heuristics \cite{stumberg:2016}, identifying the frontiers between known and unknown parts of the scene \cite{heng:2014,shen:2012}, maximizing newly visible parts of the scene \cite{bircher:2016}, maximizing information gain \cite{charrow:2015b,charrow:2015a}, and imitation learning \cite{choudhury:2017}.
A closely related problem in robotics is generating trajectories that \emph{cover} a known environment \cite{galceran:2013}.
Several coverage path planning algorithms have been proposed for drones \cite{alexis:2015,bircher:2015,heng:2015,hollinger:2013}.
In an especially similar spirit to our work, Heng et al.~propose to reconstruct an unknown environment by executing alternating exploration and coverage trajectories \cite{heng:2015}.
However, existing strategies for exploration and coverage do not explicitly account for the domain-specific requirements of multi-view stereo algorithms (e.g., observing the scene geometry from a diverse set of viewing angles).
Moreover, existing exploration and coverage strategies have not been shown to produce visually pleasing multi-view stereo reconstructions, and are generally not evaluated on multi-view stereo reconstruction tasks.
In contrast, our trajectories cover the scene in a way that explicitly accounts for the requirements of multi-view stereo algorithms, and we evaluate the multi-view stereo reconstruction performance of our algorithm directly.

Several path planning algorithms have been proposed for drones, that explicitly attempt to maximize multi-view stereo reconstruction performance \cite{dunn:2009a,hoppe:2012,mostegel:2016,schmid:2012}.
These algorithms are similar in spirit to ours, but adopt a two phase strategy for generating trajectories.
In the first phase, these algorithms select a sequence of \emph{next-best-views} to visit, ignoring travel costs.
In the second phase, they find an efficient path that connects the previously selected views.
In contrast, our algorithm reasons about these two problems -- selecting good views and routing between them -- jointly in a unified global optimization problem, enabling us to generate more rewarding trajectories, and ultimately higher-quality 3D\ reconstructions.

\vspace{-13pt}
\paragraph{View Selection and Path Planning}
The problem of optimizing the placement (and motion) of sensors to improve performance on a perception task is a classical problem in computer vision and robotics, where it generally goes by the name of \emph{active vision}, e.g., see the comprehensive surveys \cite{chen:2011,scott:2003,tarabanis:1995}.
We discuss directly related work not included in these surveys here.
A variety of active algorithms for 3D scanning with ground-based range scanners have been proposed, that select a sequence of next-best-views \cite{krainin:2011}, and then find an efficient path to connect the views \cite{fan:2016,wu:2014}.
In a similar spirit to our work, Wang et al.~propose a unified optimization problem that selects rewarding views, while softly penalizing travel costs \cite{wang:2007}.
We adapt these ideas to account for the domain-specific requirements of multi-view stereo algorithms, and we impose a hard travel budget constraint, which is an important safety requirement when designing drone trajectories.

Several algorithms have been proposed to select an appropriate subset of views for multi-view stereo reconstruction \cite{dunn:2009b,hornung:2008,mauro:2014b,mauro:2014a}, and to optimize coverage of a scene \cite{ghanem:2015,mavrinac:2013}.
However, these methods do not model travel costs between views.
In contrast, we impose a hard constraint on the travel cost of the path formed by the views we select.

\vspace{-14pt}
\paragraph{Submodular Path Planning}
Submodularity \cite{krause:2014} has been considered in path planning scenarios before, first in the theory community \cite{chekuri:2012,chekuri:2005}, and more recently in the artificial intelligence \cite{singh:2009a,singh:2009b,zhang:2016} and robotics \cite{heng:2015,hollinger:2013} communities.
The coverage path planning formulation of Heng et al.~\cite{heng:2015} is similar to ours, in the sense that both formulations use the same technique for approximating coverage \cite{iyer:2013b,iyer:2013a}.
We extend this formulation to account for the domain-specific requirements of multi-view stereo algorithms, and we evaluate the multi-view stereo reconstruction performance of our algorithm directly.

%% file: 30_approach.tex
\vspace{-5pt}
\section{Technical Overview}
\label{sec:overview}

In order to generate scanning trajectories, our algorithm leverages a coarse estimate of the scene geometry.
Initially, we do not have any estimate of the scene geometry, so we adopt an \emph{explore-then-exploit} approach.

In the \emph{explore} phase, we fly our drone (i.e., we command our drone to fly autonomously) along a default trajectory at a safe distance above the scene, acquiring a sequence of images as we are flying.
We land our drone, and subsequently feed the acquired images to an open-source multi-view stereo pipeline, thereby obtaining a coarse estimate of the scene geometry, and  a strictly conservative estimate of the scene's free space.
We include a more detailed discussion of our \emph{explore} phase in the supplementary material.

In the \emph{exploit} phase, we use this additional information about the scene to plan a scanning trajectory that attempts to maximize the fidelity of the resulting 3D reconstruction.
At the core of our planning algorithm, is a coverage model that accounts for the domain-specific requirements of multi-view stereo reconstruction (Section \ref{sec:coverage_model}).
Using this model, we generate a scanning trajectory that maximizes scene coverage, while respecting the drone's limited flight time (Section \ref{sec:trajectories}). 
We fly the drone along our scanning trajectory, acquiring another sequence of images.
Finally, we land our drone again, and we feed all the images we have acquired to our multi-view stereo pipeline to obtain a detailed 3D reconstruction of the scene.

\begin{figure}[t]
\begin{center}
\includegraphics[width=0.47\textwidth]{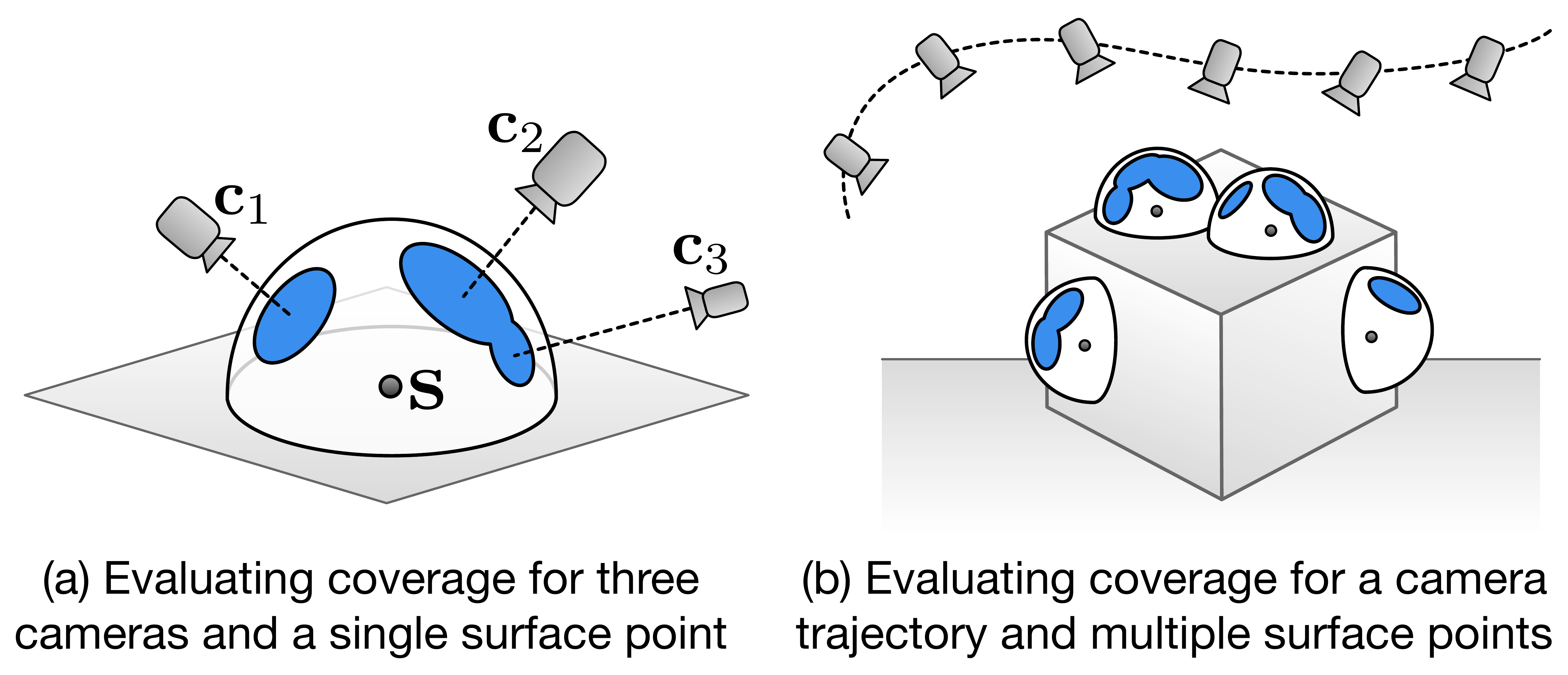}{\vspace{-7pt}}
\end{center}
\caption{
Our coverage model for quantifying the usefulness of camera trajectories for multi-view stereo reconstruction. More useful trajectories cover more of the hemisphere of viewing angles around surface points.
(a) An illustrative example showing coverage of a single surface point with three cameras. Each camera covers a circular disk on a hemisphere around the surface point $\mathbf{s}$, and the total solid angle covered by all the disks determines the total usefulness of the cameras.
Note that the angular separation (i.e., baseline) between cameras $\mathbf{c}_2$ and $\mathbf{c}_3$ is small and leads to diminishing returns in their combined usefulness.
(b) The usefulness of a camera trajectory, integrated over multiple surface points, is determined by summing the total covered solid angle for each of the individual surface points. Our model naturally encourages diverse observations of the scene geometry, and encodes the eventual diminishing returns of additional observations.
\vspace{-10pt}
}
\label{fig:coverage_model}
\end{figure}
\begin{figure*}[t]
\begin{center}
\includegraphics[width=1.0\textwidth]{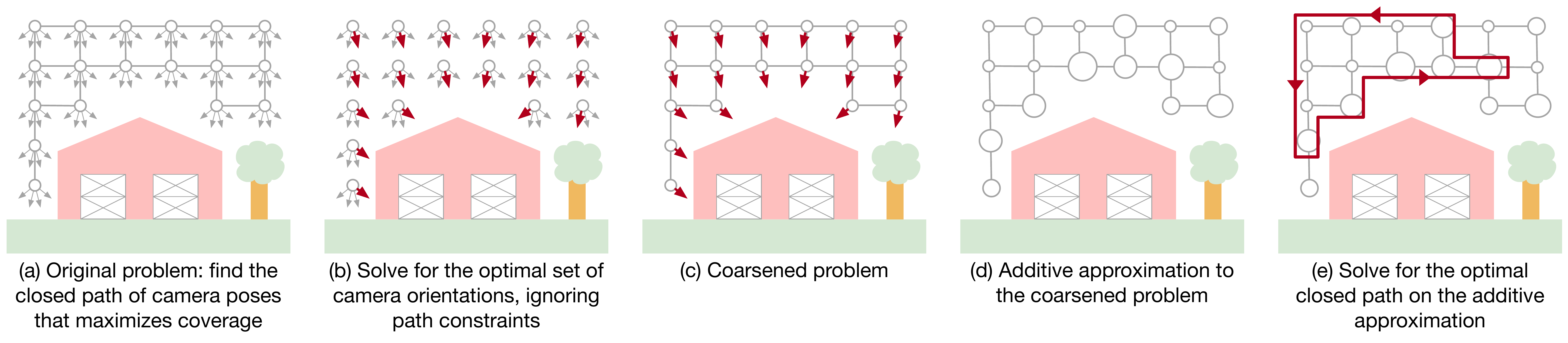}{\vspace{-10pt}}
\end{center}
\caption{
Overview of our algorithm for generating camera trajectories that maximize coverage.
(a) Our goal is to find the optimal closed path of camera poses through a discrete graph.
(b) We begin by solving for the optimal camera orientation at every node in our graph, ignoring path constraints.
(c) In doing so, we remove the choice of camera orientation from our problem, coarsening our problem into a more standard form.
(d) The solution to the problem in (b) defines an approximation to our coarsened problem, where there is an additive reward for visiting each node.
(e) Finally, we solve for the optimal closed path on the additive approximation defined in (d).
\vspace{-10pt}
}
\label{fig:algorithm_overview}
\end{figure*}

\vspace{-5pt}
\section{Coverage Model for Camera Trajectories}
\label{sec:coverage_model}

In this section, we model the usefulness of a camera trajectory for multi-view stereo reconstruction, in terms of how well it covers the scene geometry.
We provide an overview of our coverage model in Figure \ref{fig:coverage_model}.

In reality, the most useful camera trajectory is the one that yields the highest-quality 3D reconstruction of the scene.
However, it is not clear how we would search for such a camera trajectory directly, without resorting to flying candidate trajectories and performing expensive 3D reconstructions for each of them.
In contrast, our coverage model only roughly approximates the true usefulness of a camera trajectory. 
However, as we will see in the following section, our coverage model: (1) is motivated by established best practices for multi-view stereo image acquisition; (2) is easy to evaluate; (3) only requires a coarse estimate of the scene geometry as input;  and (4) exhibits submodular structure, which will enable us to efficiently maximize it.

\vspace{-10pt}
\paragraph{Best Practices for Multi-View Stereo Image Acquisition}
As a rule of thumb, it is recommended to capture an image every 5--15 degrees around an object, and it is generally accepted that capturing images more densely will eventually lead to diminishing returns in the fidelity of the 3D reconstruction \cite{furukawa:2015}.
Similarly, close-up and fronto-parallel views can help to resolve fine geometric details, because these views increase the effective resolution of estimated depth images, and contribute more reliable texture information to the reconstruction \cite{waechter:2014}.
We explicitly encode these best practices for multi-view stereo image acquisition into our coverage model.

\vspace{-10pt}
\paragraph{Formal Definition}
Given a candidate camera trajectory and approximate scene geometry as a triangle mesh, our goal is to quantify how well the trajectory covers the scene geometry.
We first uniformly sample the camera trajectory to generate a discrete set $C$, consisting of individual camera poses $\mathbf{c}_{0:I}$.
Similarly, we uniformly sample oriented surface points $\mathbf{s}_{0:J}$ from the scene geometry.
For each oriented surface point $\mathbf{s}_j$, we define an oriented hemisphere $H_j$ around it.
For each surface point $\mathbf{s}_j$ and camera $\mathbf{c}_i$, we define a circular disk $D^j_i$ that covers an angular region of the hemisphere $H_j$, centered at the location where $\mathbf{c}_i$ projects onto $H_j$ (see Figure \ref{fig:coverage_model}).
When the surface point $\mathbf{s}_j$ is not visible from the camera $\mathbf{c}_i$, we define the disk $D^j_i$ to have zero radius, and we truncate the extent of each disk so that it does not extend past the equator of $H_j$.
We define the total covered region of the hemisphere $H_j$ as the union of all the disks that partially cover $H_j$ (see Figure \ref{fig:coverage_model}), referring to this total covered region as $V_j = \bigcup_{i=0}^I D^j_i$.
We define our coverage model as follows,
\begin{equation}
\begin{aligned}
f(C) = \sum_{j=0}^{J} \int_{V_j} w_j(\mathbf{h}) d\mathbf{h}
\end{aligned}
\label{eqn:model_irregular}
\end{equation}
where the outer summation is over all hemispheres; $\int_{V_j} d\mathbf{h}$ refers to the surface integral over the covered region $V_j$; and $w_j(\mathbf{h})$ is a non-negative weight function that assigns different reward values for covering different parts of $H_j$.
Our model can be interpreted as quantifying how well a set of cameras covers the scene's \emph{surface light field} \cite{davis:2012,wood:2000}.
We include a method for efficiently evaluating our coverage model in the supplementary material.

To encourage close-up views, we set the radius of $D^j_i$ to decay exponentially as the camera $\mathbf{c}_i$ moves away from the surface point $\mathbf{s}_j$. To encourage fronto-parallel views, we design each function $w_j(\mathbf{h})$ to decay in a cosine-weighted fashion, as the hemisphere location $\mathbf{h}$ moves away from the hemisphere pole.
We include our exact formulation for $D^j_i$ and $w_j(\mathbf{h})$ in the supplementary material.

\vspace{-10pt}
\paragraph{Submodularity}
Roughly speaking, a set function is submodular if the marginal reward for adding an element to the input set always decreases, as more elements are added to the input set \cite{krause:2014}.
Our coverage model is submodular, because all coverage functions with non-negative weights are submodular \cite{krause:2014}.
Submodularity is a useful property to identify when attempting to optimize a set function, and is often referred to as the discrete analogue of convexity.
We will leverage submodularity extensively in the following section, as we derive our algorithm for generating camera trajectories that maximizing coverage. 

\vspace{-5pt}
\section{Generating Optimal Camera Trajectories}
\label{sec:trajectories}
We provide an overview of our algorithm in Figure \ref{fig:algorithm_overview}.
Our approach is to formulate a reward-collecting optimization problem on a graph.
The nodes in the graph represent camera positions, the edges represent Euclidean distances between camera positions, and the rewards are collected by visiting new nodes.
The goal is to find a path that collects as much reward as possible, subject to a budget constraint on the total path length.
This general problem is known as the \emph{orienteering problem} \cite{gunawan:2016,vansteenwegena:2011}.

A variety of approaches have been proposed to approximately solve the orienteering problem, which is NP-hard.
However, these methods are not directly applicable to our problem, because they assume that the rewards on nodes are additive.
But the total reward we collect in our problem is determined by our coverage model, which does not exhibit additive structure.
Indeed, the marginal reward we collect at a node might be very large, or very small, depending on the entire set of other nodes we visit.

The marginal reward we collect at each node also depends strongly on the orientation of our camera. In other words, our orienteering problem involves extra choices -- how to orient the camera at each visited node -- and these choices are globally coupled through our submodular coverage function. Therefore, even existing algorithms for \emph{submodular orienteering} \cite{chekuri:2012,chekuri:2005,heng:2015,singh:2009a,singh:2009b,zhang:2016} are not directly applicable to our problem, because these algorithms assume there are no extra choices to make at each visited node.

Our strategy will be to apply two successive problem transformations.
First, we leverage submodularity to solve for the approximately optimal camera orientation at every node in our graph, ignoring path constraints (Fig.~\ref{fig:algorithm_overview}b, Section \ref{sec:trajectories_orientation}).
In doing so, we remove the choice of camera orientation from our orienteering problem, thereby coarsening it into a more standard form (Fig.~\ref{fig:algorithm_overview}c).
Second, we leverage submodularity to construct a tight additive approximation of our coverage function (Fig.~\ref{fig:algorithm_overview}d, Section \ref{sec:trajectories_subgradient}).
In doing so, we relax our coarsened submodular orienteering problem into a standard additive orienteering problem.
We formulate this additive orienteering problem as a compact integer linear program, and solve it approximately using a commercially available solver (Fig.~\ref{fig:algorithm_overview}e, Section \ref{sec:trajectories_ilp}).

\vspace{-10pt}
\paragraph{Preprocessing}
We begin by constructing a discrete set of all the possible camera poses we might include in our path.
We refer to this set as our \emph{ground set} of camera poses, $C$.
We construct this set by uniformly sampling a user-defined bounding box that spans the scene, then uniformly sampling a downward-facing unit hemisphere to produce a set of look-at vectors that our drone camera can actuate. We define our ground set as the Cartesian product of these positions and look-at vectors.
We construct the graph for our orienteering problem as the grid graph of all the unique camera positions in $C$, pruned so that it is entirely restricted to the known free space in the scene  (see Section \ref{sec:overview}).

\vspace{-10pt}
\paragraph{Our Submodular Orienteering Problem}
Let $\mathbf{P} = (\mathbf{p}_0, \mathbf{v}_0), (\mathbf{p}_1, \mathbf{v}_1), \ldots, (\mathbf{p}_q, \mathbf{v}_q)$ be a camera path through our graph, represented as a sequence of camera poses taken from our ground set.
We represent each camera pose as a position $\mathbf{p}_i$ and a look-at vector $\mathbf{v}_i$.
We would like to find the optimal path $\mathbf{P}^{\star}$ as follows,
\begin{equation}
\begin{aligned}
\mathbf{P}^{\star} = \argmax_{\mathbf{P}} f(C_{\mathbf{P}})~~~~~~~~~~~~~~~~\\
\text{subject to} ~~~~~ l(\mathbf{P}) \leq B ~~~~~ \mathbf{p}_0 = \mathbf{p}_q = \mathbf{p}_\text{root}
\end{aligned}
\label{eqn:submodular_orienteering}
\end{equation}
where
$C_\mathbf{P} \subseteq C$ is the set of all the unique camera poses along the path; 
$l(\mathbf{P})$ is the length of the path; $B$ is a user-defined travel budget; and $\mathbf{p}_\text{root}$ is the position where our path must start and end.
For safety reasons, we would also like to design trajectories that consume close to, but no more than, some fixed fraction of our drone's battery (e.g., 80\% or so). However, constraining battery consumption directly is difficult to express in our orienteering formulation, so we model this constraint indirectly by imposing a budget constraint on path length.
We make the observation that our problem is intractable in its current form, because it requires searching over an exponential number of paths through our graph. This observation motivates the following two problem transformations.

\subsection{Solving for Optimal Camera Orientations} 
\label{sec:trajectories_orientation}

Our goal in this subsection is to solve for the optimal camera orientation at every node in our graph, ignoring path constraints. 
We achieve this goal with the following relaxation of the problem in equation (\ref{eqn:submodular_orienteering}).
Let $C_S \subseteq C$ be a subset of camera poses from our ground set. We would like to find the optimal subset of camera poses  $C_S^{\star}$ as follows,
\begin{equation}
\begin{aligned}
C_S^{\star} = \argmax_{C_S} f(C_S)~~~~~~~~~\\
\text{subject to} ~~~~~ \left\vert C_S \right\vert = N ~~~~~ C_S \in \mathcal{M}
\end{aligned}
\label{eqn:cardinality_matroid}
\end{equation}
where
$\left\vert C_S \right\vert$ is the cardinality of $C_S$;
$N$ is the total number of unique positions in our graph;
and the constraint $C_S \in \mathcal{M}$ enforces mutual exclusion, where we are allowed to select at most one camera orientation at each node in our graph.
In this relaxed problem, we are attempting to maximize coverage by selecting exactly one camera orientation at each node in our graph.
We can interpret such a solution as a coarsened ground set for the problem in equation (\ref{eqn:submodular_orienteering}), thereby transforming it into a standard submodular orienteering problem.

Because our coverage function is submodular, the problem in equation (\ref{eqn:cardinality_matroid}) can be solved very efficiently, and to within 50\% of global optimality, with a very simple greedy algorithm \cite{krause:2014}.
Roughly speaking, the greedy algorithm selects camera poses from our ground set in order of marginal reward, taking care to respect the mutual exclusion constraint, until no more elements can be selected.
Submodularity can also be exploited to significantly reduce the computation time required by the greedy algorithm (e.g., from multiple hours to a couple of minutes, for the problems we consider in this paper) \cite{krause:2014}.
The approximation guarantee in this subsection relies on the fact that selecting more camera poses never reduces coverage, i.e., our coverage function exhibits a property known as \emph{monotonicity} \cite{krause:2014}.
We include a more detailed discussion of the greedy algorithm, and provide pseudocode, in the supplementary material.

\subsection{Additive Approximation of Coverage}
\label{sec:trajectories_subgradient}

Our goal in this subsection is to construct an additive approximation of coverage.
In other words, we would like to define an additive reward at each node in our graph, that closely approximates our coverage function for arbitrary subsets of visited nodes.

To construct our additive approximation, we draw inspiration from the approach of Iyer et al.~\cite{iyer:2013b,iyer:2013a}.
We begin by choosing a permutation of elements in our coarsened ground set.
Let $\mathbf{C} = \mathbf{c}_0, \mathbf{c}_1, \ldots, \mathbf{c}_N$ be our permutation, where $\mathbf{c}_i$ is the $i^{\text{th}}$ element of our coarsened ground set in permuted order.
Let $C_i = \{ \mathbf{c}_0, \mathbf{c}_1, \ldots, \mathbf{c}_{i-1} \}$ be the subset containing the first $i$ elements of our permutation.
We define the additive reward for each element in our permutation as $\tilde{f}_i = f(C_i \cup \mathbf{c}_i) - f(C_i)$.
For an arbitrary subset $C_S$, our additive approximation is simply the sum of additive rewards for each element in $C_S$. Due to submodularity, this additive approximation is guaranteed to be exact for all subsets $C_i$, and to underestimate our true coverage function for all other subsets.
This guarantee is useful for our purposes, because any solution we get from optimizing our additive approximation will yield an equal or greater reward on our true coverage function.

When choosing a permutation, it is generally advantageous to place camera poses with the greatest marginal reward at the front of our permutation.
With this intuition in mind, we form our permutation by sorting the camera poses in our coarsened ground set according to their marginal reward.
Fortunately, we have already computed this ordering in Section \ref{sec:trajectories_orientation} using the greedy algorithm.
So, we simply reuse this ordering to construct our additive approximation.

\subsection{Orienteering as an Integer Linear Program}
\label{sec:trajectories_ilp}

After constructing our additive approximation of coverage, we obtain the following additive orienteering problem, 
\begin{equation}
\begin{aligned}
\mathbf{P}^{\star} = \argmax_{\mathbf{P}} \sum_{C_{\mathbf{P}}} \tilde f_i~~~~~~~~~~~~~~~\\
\text{subject to} ~~~~~ l(\mathbf{P}) \leq B ~~~~~ \mathbf{p}_0 = \mathbf{p}_q = \mathbf{p}_\text{root}
\end{aligned}
\label{eqn:ilp}
\end{equation}
where $\tilde f_i$ is the additive reward for each unique      node along the path $\mathbf{P}$.
In its current form, it is still not clear how to solve this problem efficiently, because we must still search over an exponential number of paths through our graph.
Fortunately, we can express this problem as a compact integer linear program, using a formulation suggested by Letchford et al.~\cite{letchford:2013}.
We transform our undirected graph into a directed graph, and we define integer variables to represent if nodes are visited and directed edges are traversed.
Remarkably, we can constrain the configuration of these integer variables to form only valid paths through our graph, with a compact set of linear constraints.
We include a more detailed derivation of this formulation in the supplementary material.

Leveraging the formulation suggested by Letchford et al., we convert the problem in equation (\ref{eqn:ilp}) into a standard form that can be given directly to an off-the-shelf solver.
We use the modeling language CVXPY \cite{cvxpy:2016} to specify our problem, and we use the commercially available Gurobi Optimizer \cite{gurobi:2017} as the back-end solver.
Solving integer programming problems to global optimality is NP-hard, and can take a very long time, so we specify a solver time limit of 5 minutes.
Gurobi returns the best feasible solution it finds within the time limit, along with a worst-case optimality gap.
In our experience, Gurobi consistently converges to a close-to-optimal solution in the allotted time (i.e., typically with an optimality gap of less than 10\%).
At this point, the resulting orienteering trajectory can be safely and autonomously executed on our drone.

%% file: 40_results.tex
\begin{figure*}[t]
\begin{center}
\includegraphics[width=0.98\textwidth]{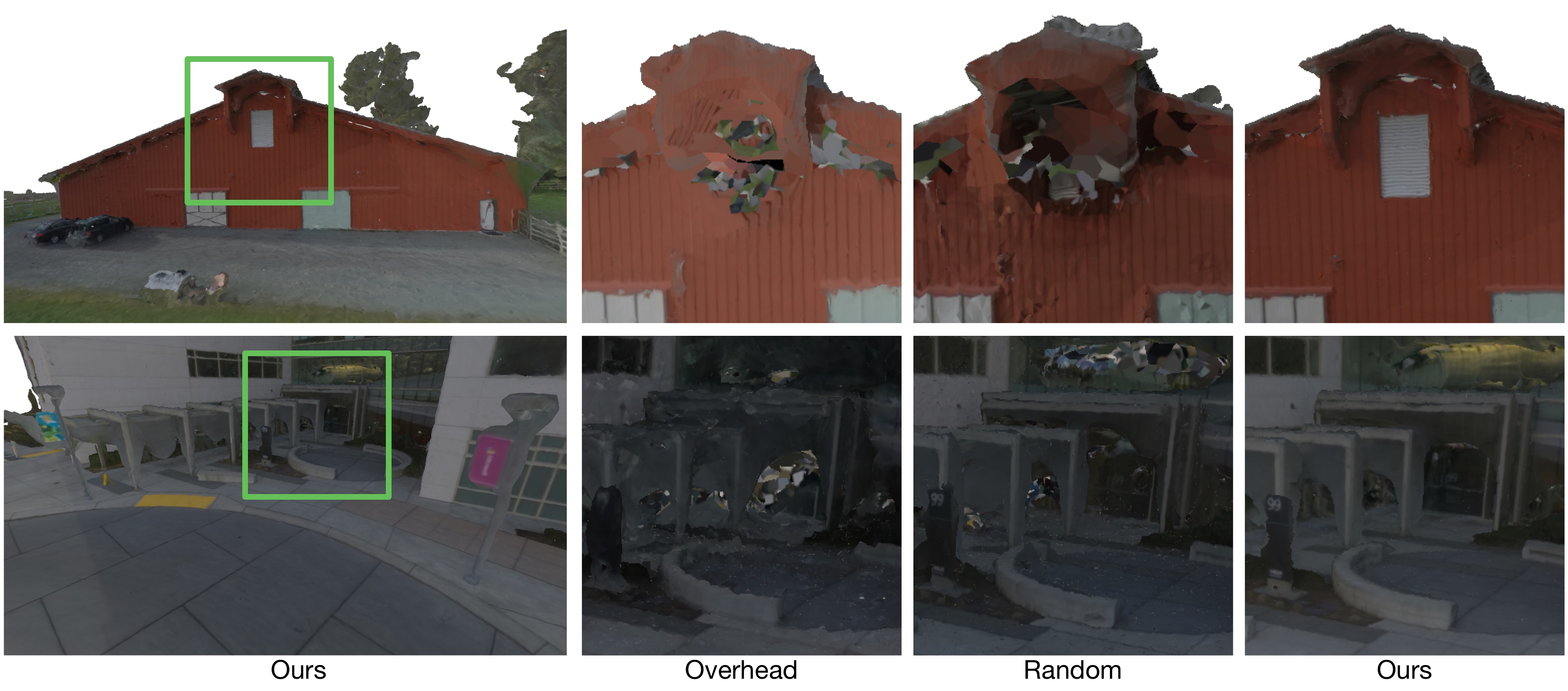}{\vspace{-7pt}}
\end{center}
\caption{
Qualitative comparison of the 3D reconstructions obtained from an overhead trajectory, a random trajectory, and our trajectory for two real-world scenes.
Our reconstructions contain noticeably fewer visual artifacts than the baseline reconstructions.
In all our experiments, we control for the flight time, battery consumption, number of images, and quality settings used in the 3D reconstruction.
\vspace{-2pt}
}
\label{fig:results_side_by_side}
\end{figure*}

\begin{figure*}[t]
\begin{center}
\includegraphics[width=0.98\textwidth]{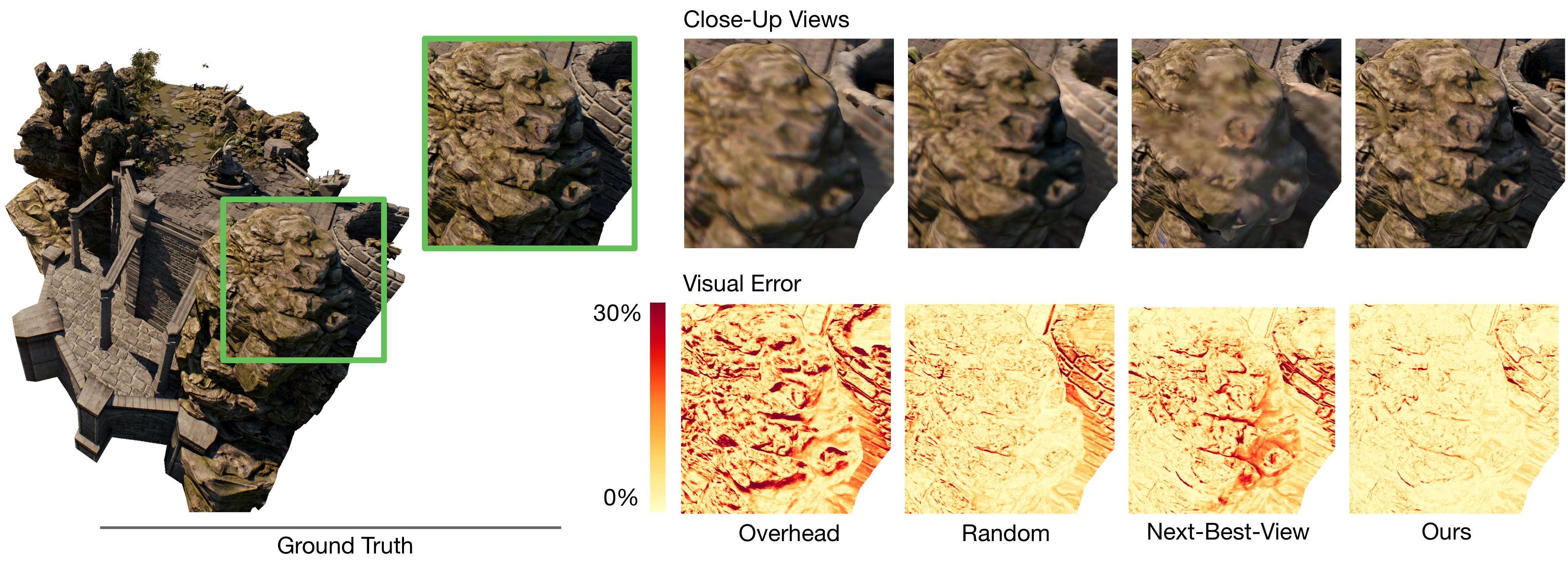}{\vspace{-7pt}}
\end{center}
\caption{
Quantitative comparison of the 3D reconstructions obtained from an overhead trajectory, a random trajectory, a next-best-view trajectory, and our trajectory for our synthetic scene.
We show close-up renderings of each reconstruction, as well as per-pixel visual error, relative to a ground truth rendering of the scene.
Our method leads to quantitatively lower visual error than baseline methods.
\vspace{-12pt}
}
\label{fig:results_quantitative}
\end{figure*}

\vspace{-5pt}
\section{Evaluation}

\vspace{-1pt}
In all the experiments described in this section, we execute all drone flights at 2 meters per second, with a total travel budget of 960 meters (i.e., an 8 minute flight) unless otherwise noted.
All flights generate 1 image every 3.5 meters.
Each method has the same travel budget, and generates roughly 275 images.
Small variations in the number of generated images are possible, due to differences in how close each method gets to the travel budget.
We describe our drone hardware, data acquisition pipeline, and experimental methodology in more detail in the supplementary material.

\vspace{3pt}
\textbf{Real-World Reconstruction Performance}
We evaluated the real-world reconstruction performance of our algorithm by using it to scan three large outdoor scenes: a barn, an office building, and an industrial site.\footnote{We
conducted this experiment with an early implementation of our method that differs slightly from the implementation used in our other experiments.
In particular, the graph of camera positions used in this experiment included diagonal edges.
We subsequently excluded diagonal edges to enable our integer programming formulation to scale to larger problem instances. 
}
We show results from these experiments in Figures \ref{fig:teaser} and \ref{fig:results_side_by_side}, as well as in the supplementary material.
We compared our reconstruction results to two baseline methods: \textsc{Overhead} and \textsc{Random}.

\textsc{Overhead}.
We designed \textsc{Overhead} to generate trajectories that are representative of those produced by existing commercial flight planning software \cite{3dr:2017a,pix4d:2017a}.
\textsc{Overhead} generates a single flight at at a safe height above the scene; consisting of an orbit path that always points the camera at the center of the scene; followed by a lawnmower path that always points the camera straight down.

\textsc{Random}.
We designed \textsc{Random} to have roughly the same level of scene understanding as our algorithm, except that \textsc{Random} does not optimize our coverage function.
We gave \textsc{Random} access to the graph of camera positions generated by our algorithm, which had been pruned according to the free space in the scene.
\textsc{Random} generates trajectories by randomly selecting graph nodes to visit and traveling to them via shortest paths, until no more nodes can be visited due to the travel budget.
\textsc{Random} always points the camera towards the center of the scene, which is a reasonable strategy for the scenes we consider in this paper.

During our \emph{explore} phase, we generate an orbit trajectory exactly as we do for \textsc{Overhead}.
For the scenes we consider in this paper, this initial orbit trajectory is always less than 250 meters.

When generating 3D\ reconstructions, our algorithm and \textsc{Random} have access to the images from our \emph{explore} phase, but \textsc{Overhead} does not.
The images in our \emph{explore} phase are nearly identical to the orbit images from \textsc{Overhead}, and would therefore provide \textsc{Overhead} with negligible additional information, so all three methods are directly comparable.
We generated 3D reconstructions using the commercially available Pix4Dmapper Pro software \cite{pix4d:2017b}, configured with maximum quality settings.

\vspace{-1pt}
\textbf{Reconstruction Performance on a Synthetic Scene}
We evaluated our algorithm using a photorealistic video game simulator, which
enabled us to measure reconstruction performance relative to known ground truth geometry and appearance.
We show results from this experiment in Figure \ref{fig:results_quantitative} and Table \ref{tbl:quantitative}.

Our experimental design here is exactly as described previously, except we acquired images by programmatically maneuvering a virtual camera in the Unreal Engine \cite{epic:2017a}, using the UnrealCV Python library \cite{qiu:2016}.
We also included an additional baseline method, \textsc{Next-Best-View}, that greedily selects nodes according to their marginal submodular reward, and finds an efficient path to connect them using the Approx-TSP algorithm \cite{cormen:2009} until no more nodes can be added due to the travel budget.
This method is intended to be representative of the \emph{next-best-view} planning strategies that occur frequently in the literature \cite{fan:2016,hollinger:2013,krainin:2011,wu:2014}, including those that have been applied to aerial 3D scanning \cite{dunn:2009a,hoppe:2012,mostegel:2016,schmid:2012}.

We chose the \textsc{Grass Lands} environment \cite{epic:2017b} as our synthetic test scene because it is freely available, has photorealistic lighting and very detailed geometry, and depicts a large outdoor scene that would be well-suited for 3D scanning with a drone.

We evaluated geometric reconstruction quality by measuring \emph{accuracy} and \emph{completeness} relative to a ground truth point cloud \cite{aanaes:2016,knapitsch:2017}.
We obtained our ground truth point cloud by rendering reference depth images arranged on an inward-looking sphere around the scene, taking care to manually remove any depth images that were inside objects.
We also evaluated visual reconstruction quality by measuring \emph{per-pixel visual error}, relative to ground truth RGB images rendered from the same inward-looking sphere around the scene \cite{waechter:2017}.
When evaluating per-pixel visual error, we took care to only compare pixels that contain geometry from inside the scanning region-of-interest for our scene.

When evaluating geometric quality, we obtained point clouds for each method by running VisualSFM \cite{wu:2013,wu:2007,wu:2011b,wu:2011a}, followed by the Multi-View Environment \cite{fuhrmann:2015}, followed by Screened Poisson Surface Reconstruction \cite{kazhdan:2013}, and finally by uniformly sampling points on the reconstructed triangle mesh surface.
When evaluating visual quality, we obtained textured 3D models for each method using the surface texturing algorithm of Waechter et al.~\cite{waechter:2014}.

\begin{table}[t]
\centering
\footnotesize
\begin{tabular}{@{}llll@{}}
\toprule
Method         & Accuracy       & Completeness   & Visual     \\
               & Error (mm)     & Error (mm)     & Error (\%) \\
\midrule
Overhead       & 170.2          & 583.8          & 7.1  \\
Random         & 126.5          & 557.2          & 4.4 \\
Next-Best-View & 122.8          & 330.7          & 3.6 \\
\textbf{Ours}  & \textbf{115.2} & \textbf{323.3} & \textbf{3.3} \\
\bottomrule
\end{tabular}
\normalsize
\vspace{5pt}
\caption{
Quantitative comparison of the 3D reconstructions obtained from an overhead trajectory, a random trajectory, a next-best-view trajectory, and our trajectory for our synthetic scene.
For all the columns in this table, lower is better.
We report the mean per-pixel visual error across all of our test views, where 100\% per-pixel error corresponds to the $l_2$ norm of the difference between black and white in RGB space.
Our method quantitatively outperforms baseline methods, both geometrically (i.e., in terms of accuracy and completeness) and visually.
\vspace{-13pt}
}
\label{tbl:quantitative}
\end{table}

\begin{figure}[t]
\begin{center}
\includegraphics[width=0.42\textwidth]{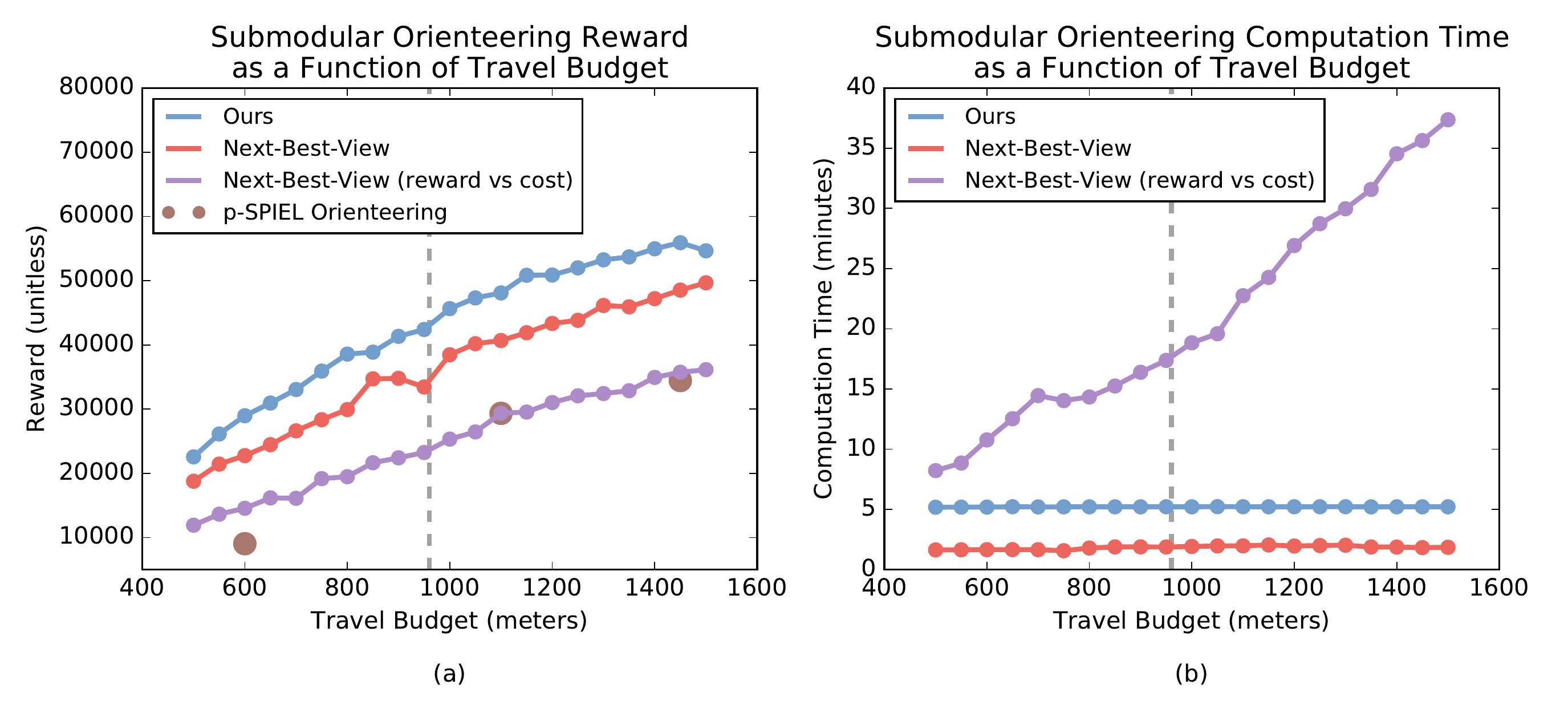}{\vspace{-10pt}}
\end{center}
\caption{
Quantitative comparison of submodular orienteering algorithms on our synthetic scene.
(a) Submodular reward as a function of travel budget.
Our algorithm consistently obtains more reward than other algorithms.
All reconstruction results in this paper were produced with a budget of 960 meters (i.e., 8 minutes at 2 meters per second), shown with a grey dotted line.
For this budget, we obtain 20\% more reward than next-best-view planning. The p-SPIEL Orienteering algorithm \cite{singh:2009b} failed to consistently find a solution.
(b) Computation time as a function of travel budget.
On this plot, lower is better.
In terms of computation time, our algorithm is competitive with, but more expensive than, next-best-view planning.
We do not show computation times for the p-SPIEL Orienteering algorithm, because it took over 4 hours in all cases where it found a solution.\vspace{-6pt}
\vspace{-4pt}
}
\label{fig:results_orienteering}
\end{figure}

\vspace{3pt}
\textbf{Submodular Orienteering Performance}
\label{sec:submodperf}
We evaluated the submodular orienteering performance of our algorithm on our synthetic scene.
We performed this experiment after we have solved for the optimal camera orientation at every node in our graph, to facilitate the comparison of our algorithm to other submodular orienteering algorithms \cite{singh:2009b,zhang:2016}.
We show results from this experiment in Figure \ref{fig:results_orienteering}.

In this experiment, we included a baseline method that behaves identically to \textsc{Next-Best-View}, except it greedily selects nodes according to the ratio of marginal reward to marginal cost \cite{zhang:2016}.
We implemented all algorithms in Python, except for the p-SPIEL Orienteering algorithm \cite{singh:2009b}, where we used the MATLAB implementation provided by the authors.
We performed this experiment on a Mid 2015 Macbook Pro with a 2.8 GHz Intel Core i7 processor and 16GB of RAM.


%% file: 50_discussion.tex
\vspace{-5pt}
\section{Conclusions}

We proposed an intuitive coverage model for aerial 3D scanning, and we made the observation that our model is submodular.
We leveraged submodularity to develop a computationally efficient method for generating scanning trajectories, that reasons jointly about coverage rewards and travel costs.
We evaluated our method by using it to scan three large real-world scenes, and a scene in a photorealistic video game simulator.
We found that our method results in quantitatively higher-quality 3D reconstructions than baseline methods, both geometrically and visually.




In the future, we believe trajectory optimization and geometric reasoning will enable drones to capture the physical world with unprecedented coverage and scale.
Individual drones may soon be able to execute very efficient scanning trajectories at the limits of their dynamics, and teams of drones may soon be able to execute scanning trajectories collectively and iteratively over very large scenes.

\vspace{-5pt}
\section*{Acknowledgements}

We thank Jim Piavis and Ross Robinson for their expertise as our safety pilots;
Don Gillett for granting us permission to scan the barn scene;
3D Robotics for granting us permission to scan the industrial scene;
Weichao Qiu for his assistance with the UnrealCV Python library;
Jane E and Abe Davis for proofreading the paper;
and Okke Schrijvers for the helpful discussions.
This work was supported by a generous grant from Google.

